\documentclass[10pt,twocolumn,letterpaper]{article}

\usepackage[final]{iccv}      
\usepackage{amsthm}  
\usepackage{placeins}
\newtheorem{remark}{Remark}
\usepackage{pifont}
\usepackage{multirow}
\usepackage{soul}
\definecolor{iccvblue}{rgb}{0.21,0.49,0.74}
\usepackage[pagebackref,breaklinks,colorlinks,allcolors=iccvblue]{hyperref}

\def\paperID{006} 
\def\confName{ICCV}
\def\confYear{2025}

\title{First Place Solution to the MLCAS 2025 GWFSS Challenge:\\
The Devil is in the Detail and Minority}

\author{
Songliang~Cao \quad Tianqi~Hu \quad Hao~Lu\thanks{corresponding author}\\
National Key Laboratory of Multispectral Information Intelligent Processing Technology \\
School of Artificial Intelligence and Automation\\
Huazhong University of Science and Technology, China\\
{\tt\small songliangcao@126.com hlu@hust.edu.cn}
}

\begin{document}
\maketitle
\begin{abstract}
In this report, we present our solution during the participation of the MLCAS 2025 GWFSS Challenge. This challenge hosts a semantic segmentation competition specific to wheat plants, which requires to segment three wheat organs including the head, leaf, and stem, and another background class. In 2025, participating a segmentation competition is significantly different from that in previous years where many tricks can play important roles. Nowadays most segmentation tricks have been well integrated into existing codebases such that our naive ViT-Adapter baseline has already achieved sufficiently good performance. Hence, we believe the key to stand out among other competitors is to focus on the problem nature of wheat per se. By probing visualizations, we identify the key---the stem matters. In contrast to heads and leaves, stems exhibit fine structure and occupy only few pixels, which suffers from fragile predictions and class imbalance. Building on our baseline, we present three technical improvements tailored to stems: i) incorporating a dynamic upsampler SAPA used to enhance detail delineation; ii) leveraging semi-supervised guided distillation with stem-aware sample selection to mine the treasure beneath unlabeled data; and iii) applying a test-time scaling strategy to zoom in and segment twice the image. Despite being simple, the three improvements bring us to the first place of the competition, outperforming the second place by clear margins. Code and models will be released at \url{https://github.com/tiny-smart/gwfss25}.

\end{abstract}    
\section{Introduction}
\label{sec:intro}

The 7th International Workshop on Machine Learning for Cyber-Agricultural Systems (MLCAS 2025), in conjunction with the Computer Vision in Plant Phenotyping and Agriculture (CVPPA) workshop at International Conference on Computer Vision (ICCV 2025), host a competition called Global Wheat Full Semantic Segmentation (GWFSS). 
The competition aims to advance pixel-level understanding of wheat plants, \textit{i.e.}, segmenting the full plant components, including four classes of \texttt{leaf}, \texttt{stem}, \texttt{head}, and \texttt{background}, to comprehensively describe plant architecture, health, and development. In this report, we present our winner solution to this competition. 

\begin{figure}[!t]
    \centering
    \includegraphics[width=\linewidth]{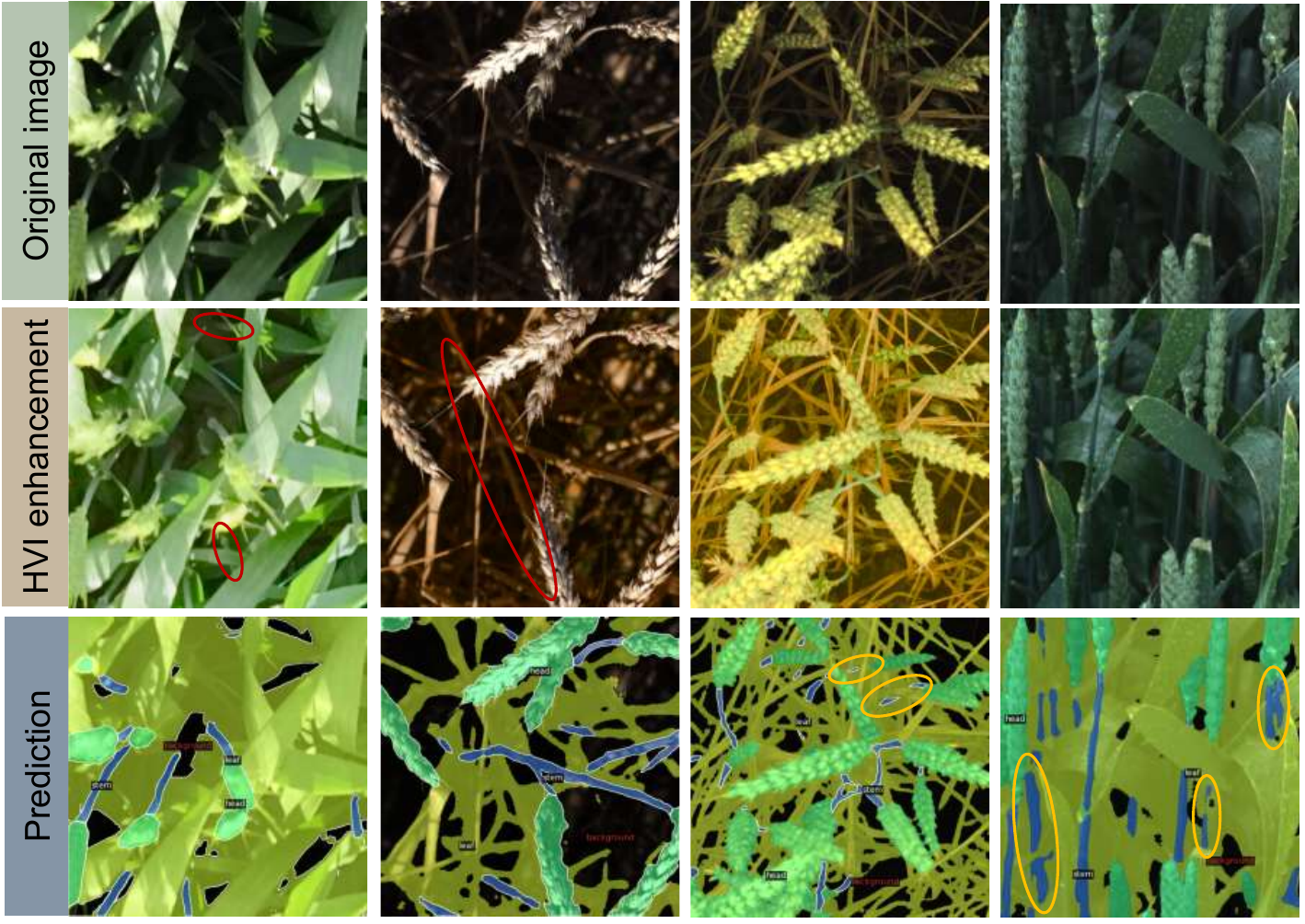}
    \caption{\textbf{Visual inspection of the ViT-Adapter predictions on the GWFSS validation set}. Our intuition tell us that the performance bottleneck seems to lie in the \texttt{stem} class, which not only requires good detail delineation but also needs to tackle the minority of pixel occupancy. Images in the second row are enhanced by the HVI color space~\cite{yan2025hvi} for ease of interpretation. The red circle seems to be stems segmented missed and the yellow one seems to be stems segmented roughly.}
    \label{fig:stage1_vis}
\end{figure}

The competition has two phases. In the development phase (Phase 1), participants can have access to a small subset of fully labeled training data (only $99$ samples) and a large amount of unlabeled data ($64,368$ samples); validation data (only images) are provided for participants to build and iterate their models. In the testing phase (Phase 2), the top 20 participants are invited to generate inference results on the test set using the final model from Phase 1. 
Different from other unconstrained open challenges, there are three rules in this competition: i) only ImageNet-1K pre-trained backbones are allowed to use; ii) only the provided GWFSS training data is allowed to use, and participants cannot use other unlabeled datatsets for pre-training; iii) manual labeling or annotation is strictly prohibited. 

To quickly gain a 
first impression of the visual challenges of this task, we trained a state-of-the-art ViT-Adapter with the Mask2Former head using the labeled training data. 
Some visualizations on the validation set are shown in the Fig.~\ref{fig:stage1_vis}. Due the significant illumination variations, we resort to a recent learnable HVI color space~\cite{yan2025hvi} for ease of interpretation of segmentation results. This color space significantly enhances the contrast of low-light regions such as shadows and reveals many missing details. According to Fig.~\ref{fig:stage1_vis}, we have the following observations and insights.

\begin{remark}
    The fine structure of the stem class leads to difficult boundary predictions such that the model needs to enhance detail awareness.
\end{remark}

By observing the patterns of visualization results, we find that 
the \texttt{head}, \texttt{leaf}, and \texttt{background} classes are well segmented in general, while the \texttt{stem} class exhibits a 
high
degree of 
fragile predictions. 
We 
therefore test the class IoU metric on the training dataset with available ground truths: the \texttt{stem} class only reports $0.69$ mIoU, while the other three classes achieve $0.88$, $0.91$, and $0.90$ for \texttt{background}, \texttt{head}, and \texttt{leaf}, respectively. It seems that the performance bottleneck lies in the \texttt{stem} class. In our view, we believe an effective way to alleviate this bottleneck is to enhance detail awareness of the model. Technically, this amounts to the question on how to improve the quality of high-resolution feature maps. 

\begin{remark}
    The model may risk overfitting due to rather limited labeled data such that unlabeled data should be exploited.
\end{remark}
The labeled training data are limited (only $99$ images are provided), but our baseline model has more than $200$M parameters, which makes it prone to overfitting. Therefore, how to effectively leveraging the unlabeled data seems to be another key to improve model generalization. In open literature, there are two mainstream ways to exploit the unlabeled data: i) resorting to self-supervised learning to pre-train a vision backbone (e.g., MAE~\cite{he2022masked} and DINOv2~\cite{oquab2023dinov2}); ii) leveraging semi-supervised learning to learn a robust model (e.g. Mean Teacher~\cite{tarvainen2017mean}). We will choose one of the two routes. 

\begin{remark}
    The stem class only occupies a small number of pixels such that the model may suffer from class imbalance.
\end{remark}
By analyzing the number of pixel labels of each class in the training dataset, we find that the proportions of \texttt{background}, \texttt{head}, \texttt{stem}, and \texttt{leaf} pixels are $85$ : $27$ : $9$ : $137$, respectively. It seems \textit{the devil is in the detail and minority}: the stem class is not only subtle to segment but also underrepresented in the dataset. 
Hence, we believe we also need to address the class imbalance issue and increase the visibility of the \texttt{stem} class during model training. 

To address the challenges above, we build upon a state-of-the-art segmentation framework ViT-Adapter~\cite{chen2022vision} and present three technical improvements. 
Considering that detail delineation is closely related to the quality of high-resolution feature maps, we first incorporate a dynamic upsampling operator SAPA~\cite{lu2022sapa} into the Mask2Former head~\cite{cheng2022masked} to enhance the detail delineation of the model. 
Second, we apply a semi-supervised learning pipeline with guided distillation~\cite{berrada2024guided} to leverage the unlabeled training set. Note that,   to alleviate the class imbalance and increase the exposure of the \texttt{stem} class, we do not use the entire unlabeled data but carefully choose a small portion of stem-related image data to form a subset (4500 images), filtered by a model trained with labeled training data. 
Finally, we conduct a form of test-time scaling with a \textit{zoom in and segment twice} strategy, which zooms in images to different scales and segment the image of each scale following a sliding-window style inference. This strategy significantly benefits the prediction of the \texttt{stem} class. 
Combining the three improvements, our final solution achieves the first place in the competition during both the development phase and the testing phase, reporting $0.77$ and $0.75$ mIoU metric, respectively, outperforming the second place by clear margins. While the three improvements seem simple, we hope our solution can inform the plant phenotyping community that a simple tweak can make a difference if one identifies the bottleneck correctly.
\section{GWFSS Competition Dataset}
The competition is based on the GWFSS dataset. The GWFSS dataset is a Siamese dataset to 
the Global Wheat Head Detection (GWHD) dataset~\cite{david2020global}. Unlike the GWHD dataset focusing on an object detection task, the GWFSS dataset highlights pixel-wise understanding, that is, semantic segmentation, of plant organs including the \texttt{head}, \texttt{stem}, and \texttt{leaf}, aiming to comprehensively describe the architecture, health, and development of wheat plants. 

\begin{figure}[!t]
    \centering
    \includegraphics[width=\linewidth]{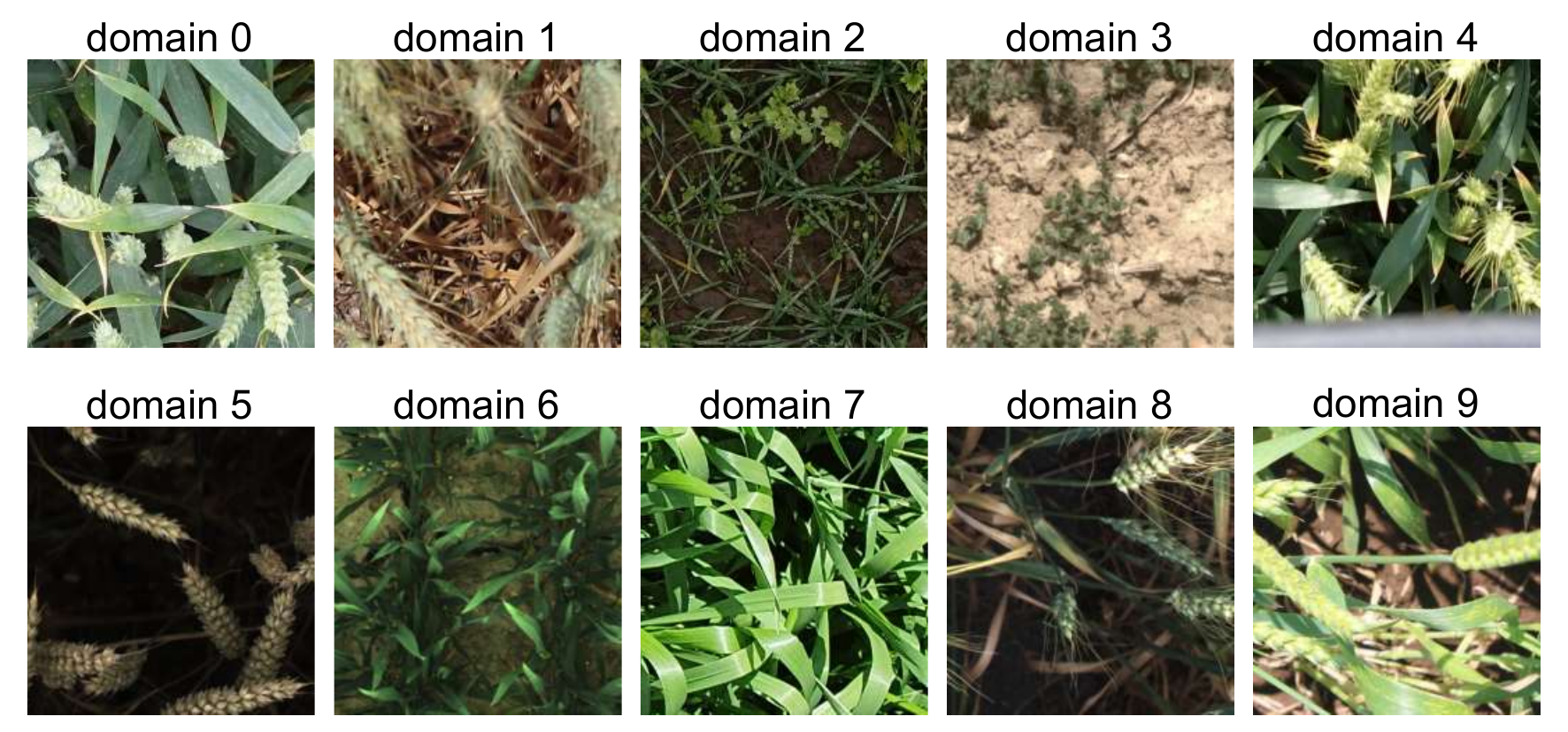}
    \caption{\textbf{Samples of different domains in the GWFSS dataset}. Domain 1-9 are used for training and validation, and domain 0 is used for testing. Different domains exhibit significant intrinsic and extrinsic variations.}
    \label{fig:dataset}
\end{figure}

Similar to the GWHD dataset, the GWFSS dataset is built with international support, with $11$ institutes and universities from diverse geographical regions using various imaging setups. Such a diversity ensures that the dataset 
covers sufficiently large variations in environmental conditions, genotypes, and illumination, which results in visual challenges such as appearance changes, shadows, and complex background. Even the wheat plants within the same domain can encompass multiple growth stages, leading to intrinsic morphological variations. As shown in Fig.~\ref{fig:dataset}, there is a substantial distribution gap between images from different domains. This renders significant difficulties in building a robust segmentation. In particular, the GWFSS dataset is split into a training set, a validation set, and a testing set. The training data and validation data are both selected from $9$ out of $10$ domains, where the labeled training data and the validation data have $99$ images, and the unlabeled training data are more than $60,000$. The left domain, with $110$ images, is used for testing data to test the generalization of the model. The data splits are summarized in Table~\ref{tab:dataset_data_num}. All images in the dataset have a resolution of $512\times 512$. Only the ground truths of the labeled training data are provided. To test on the validation data and testing data, participants must submit their results to the \texttt{codabench} online platform.\footnote{\url{https://www.codabench.org/competitions/5905/}}. 

\section{Proposed Solution}
\label{sec:method}

\begin{figure*}[!t]
    \centering
    \includegraphics[width=\linewidth]{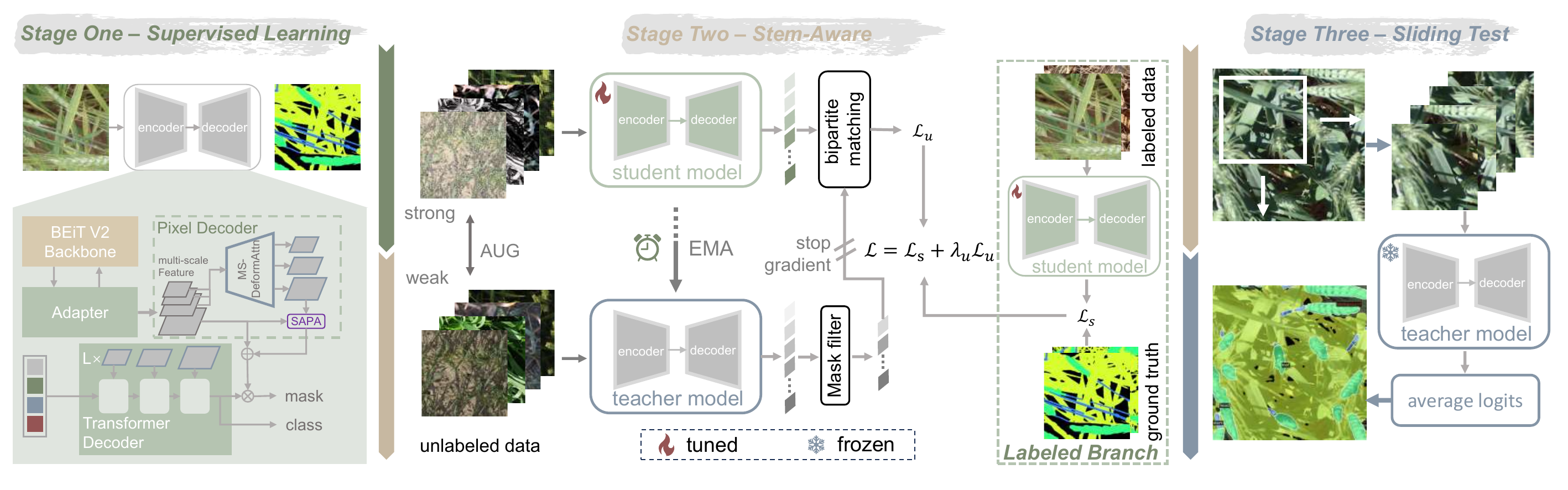}
    \caption{\textbf{The overview of our solution}. Our solution includes three stages: in stage one, we leverage the labeled training dataset to train a supervised baseline ViT-Adapter and enhance its detail delineation with a dynamic upsampler SAPA; in stage two, we apply a semi-supervised learning pipeline with guided distillation on both labeled data and selected unlabeled data; in stage three, we implement a form of test-time scaling by zooming in images and segmenting twice following the sliding-window-style inference. }
    \label{fig:pipeline}
\end{figure*}

Recently some vision foundation models such as DINOv2~\cite{oquab2023dinov2} have emerged and demonstrated robust performance across a range of downstream tasks. Indeed many top-performing solutions in open challenges are built upon fine-tuning large vision models. However, since the competition only allows to use ImageNet-1K pretrained backbones, we turn our focus to the problem nature of the task and mainly address the challenges we observe above.
The overview of our solution is 
shown in Fig.~\ref{fig:pipeline}. Our solution includes three stages: training a supervised segmentation baseline, applying a semi-supervised semantic segmentation pipeline, and leveraging a sliding-window-style test-time scaling strategy. 



\subsection{SAPA-Enhanced ViT-Adapter}

We choose the ViT-Adapter~\cite{chen2022vision} as our baseline, because current state-of-the-art semantic segmentation models are typically based on the architecture of vision transformer (ViT)~\cite{dosovitskiy2021image}. In particular, we adopt the BEiTv2 and ImageNet-1K pretrained ViT-Large model.\footnote{\url{https://github.com/microsoft/unilm/blob/master/beit2/get_started_for_image_classification.md}} 
By feeding the input image tokens into the ViT-Adapter, multi-scale features $\{\mathcal{F}_1^{\frac{1}{4}},\mathcal{F}_2^{\frac{1}{8}},\mathcal{F}_3^{\frac{1}{16}},\mathcal{F}_4^{\frac{1}{32}} \}$ are extracted, followed by the Mask2Former head~\cite{he2022masked}. In the pixel decoder of Mask2Former, the three low-resolution features $\{\mathcal{F}_2^{\frac{1}{8}},\mathcal{F}_3^{\frac{1}{16}},\mathcal{F}_4^{\frac{1}{32}}\}$ are refined through several layers of multi-scale deformable attention~\cite{xia2022vision}, and the output are denoted by \{$\mathcal{O}_2^\frac{1}{8}$, $\mathcal{O}_3^\frac{1}{16}$, $\mathcal{O}_4^\frac{1}{32}$\}. The decoder feature $\mathcal{O}_2^{\frac{1}{8}}$ is upsampled and fused with $\mathcal{F}_1^{\frac{1}{4}}$ to obtain the mask feature $\mathcal{F}_{\text{mask}}$, and $\mathcal{F}_{\text{mask}}$ is jointly used with the mask query embedding $\mathcal{Q}_{mask}$ to predict final segmentation mask $M$ as
\begin{equation}\label{eq:mask}
    M=g(\mathcal{Q}_{mask},\mathcal{F}_{mask})\,,
\end{equation}
where $g(\cdot)$ is the mask predictor.
Eq.~\eqref{eq:mask} indicates that the quality of $\mathcal{F}_{\text{mask}}$ is 
closely related to 
the quality of the final predicted masks $M$.

Recall that our goal is to enhance the prediction of the \texttt{stem} class. The fine structure of the stem requires the model to have clear detail delineation in the high-resolution mask feature $\mathcal{F}_{\text{mask}}$. According to open literature~\cite{lu2019indices,lu2022index,lu2022fade,lu2022sapa,liu2023learning,lu2025fade}, one critical stage that affects the quality of $\mathcal{F}_{\text{mask}}$ is feature upsampling. Unfortunately, the standard Mask2Former head uses bilinear interpolation for upsampling, which would \textit{ipso facto} blur details. Hence, we propose to replace bilinear upsampling with SAPA~\cite{lu2022sapa} to enhance the detail awareness of the model.

\begin{table}[!t]
    \renewcommand{\arraystretch}{1.5}
    \addtolength{\tabcolsep}{-1pt}
    \centering
    \begin{tabular}{@{}lcccc@{}}
    \toprule
        \multirow{2}*{\textbf{data splits}} & \multicolumn{2}{c}{\textbf{training}} & \multirow{2}*{\textbf{validation}} & \multirow{2}*{\textbf{testing}} \\
    \cline{2-3}
        & \textbf{labeled} & \textbf{unlabeled} & & \\
    \hline
        \#Images & $99$ & $64,368$ & $99$ & $110$   \\
    \bottomrule
    \end{tabular}
    \caption{\textbf{Training, validation, the testing splits of the GWFSS competition dataset}.}
    \label{tab:dataset_data_num}
\end{table}

SAPA 
enhances detail delineation by modeling the mutual similarity between 
the low-resolution decoder neighborhood and the high-resolution encoder feature point in the upsampling kernel. 
Formally, given a local decoder neighborhood $\mathcal{N}_l$ informed by the location $l$, the corresponding upsampling weight $w_l$ takes the form 
\begin{equation}\label{eq:kernel}
w_l = \frac{h(\text{sim}(x,y))}{\sum\limits_{z\in \mathcal{N}_l} h(\text{sim}(z,y))}\,,
\end{equation}
where $x$ is the decoder feature point, $y$ the encoder feature point, $\text{sim}(\cdot,\cdot)$ the similarity function, and $h(\cdot)$ the normalization function. It has been proved that such a formulation has conditional boundary sharpness and conditional smoothness guarantees~\cite{lu2022sapa}. We choose the similarity function to be gated similarity $\text{sim}(x,y)=gx^TPy+(1-g)x^TQy$, where $g\in(0,1)$ is a gating unit learned by linear projection, and $P$ and $Q$ are learnable projection matrices. $h(\cdot)$ is chosen to be the $\tt softmax$ normalization. In this way, Eq.~\eqref{eq:kernel} computes the bilinear kernel~\cite{tenenbaum2000separating}. With SAPA, the mask feature $\mathcal{F}_{\text{mask}}$ amounts to
\begin{equation}
  \mathcal{F}_{\text{mask}}=\mathcal{F}_1^\frac{1}{4}+\text{SAPA}(\mathcal{F}_1^\frac{1}{4}, \mathcal{O}_1^\frac{1}{8}).  
\end{equation}

\begin{figure}[!t]
    \centering
    \includegraphics[width=\linewidth]{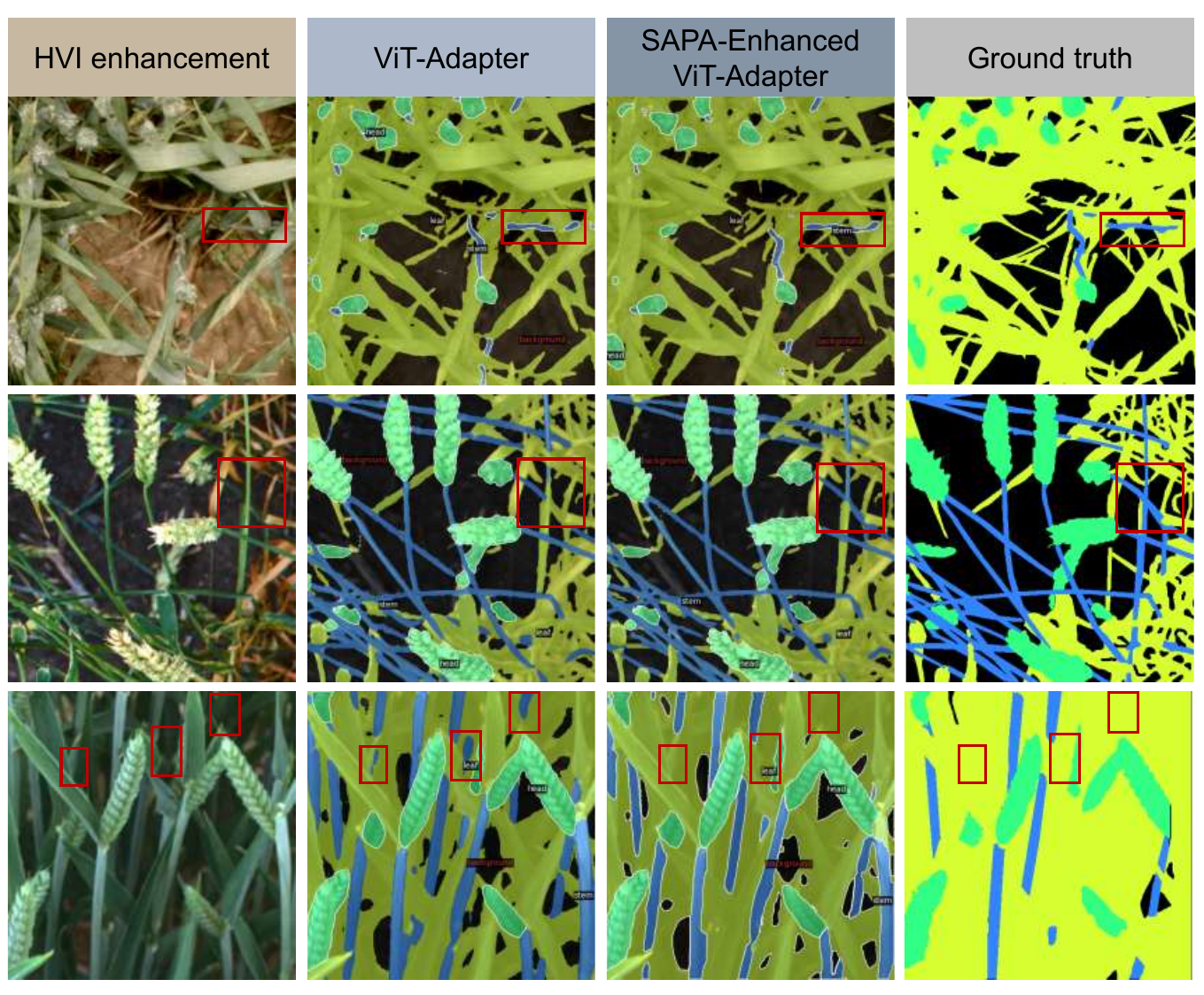}
    \caption{\textbf{Comparison between the naive ViT-Adapter baseline and SAPA-Enhanced ViT-Adapter}. The red rectangles indicate that SAPA-Enhanced ViT-Adapter is more effective than the vanilla ViT-Adapter in tackling the segmentation of stem structures, with less misclassifications and missing segmentations.}
    \label{fig:sapa_baseline}
\end{figure}

The effect of SAPA can refer to Fig.~\ref{fig:sapa_baseline}. It can be observed that SAPA effectively enhances detail delineation and reduces fragile mask predictions.

\subsection{Guided Distillation with Stem-Aware Sample Selection}
As aforementioned, 
one can choose two routes, \ie, self-supervised pretraining or semi-supervise learning, to exploit the unlabeled data provided by the competition. We finally choose the second route of semi-supervised learning. The reason is that, although there are over $60,000$ unlabeled images, which is significantly more than the $99$ labeled data, such a data scale 
may still be insufficient for self-supervised learning to train a robust vision backbone (DINOv2 uses 142M images~\cite{oquab2023dinov2}). Meanwhile, even with a well-pretrained backbone, 
fine-tuning is still required. 

Mainstream semi-supervised learning is typically based on the Mean Teacher framework~\cite{tarvainen2017mean}. In 
this framework, there is often a Siamese network 
with two identical model architectures and parameters, where a teacher model generates soft predictions to guide the training of a student model. The teacher model is updated following the exponential moving average (EMA) with the parameters of the student model. 
Let $\theta_t^{i}$ and $\theta_s^{i}$ denote the weights of the teacher model and the student model at the $i$-th iteration, respectively. 
The teacher model is updated as 
\begin{equation}
    \theta_t^{i+1}=\alpha\theta_t^{i} + (1-\alpha)\theta_s^{i+1}\,,
\end{equation}
where $\alpha$ is a damping factor.
For each unlabeled sample, the teacher and student models receive perturbed versions of the same input (e.g., different augmentations), and the student is trained to match the output of the teacher model, 
encouraging consistency between them. The meta pipeline of mean teacher enables the teacher model to learn from a large amount of unlabeled data and therefore enhances its generalization ability.

In this competition, 
we adopt an improved guided distillation framework~\cite{berrada2024guided}. The overview of guided distillation is shown in the stage two of Fig.~\ref{fig:pipeline}. 
Most existing distillation-based pipelines train a model on the labeled data as the student. However, this strategy 
can be suboptimal in our competition due to the fact that the student model 
may overfit the labeled training data (only $99$ images). To address this, guided distillation introduces a guided burn-in stage, 
where the student model is allowed to 
learn from both labeled and unlabeled data. Before guided distillation, the teacher model is first trained on the labeled data and is used to generate pseudo labels for the student model (the stage one). Then the student model is initialized randomly 
and learned from both labeled and pseudo-labeled data. After the guided burn-in stage, the student model will 
transfer its weights to the teacher model and perform the standard EMA update. 

To better adapt this framework to our problem, we further repurpose the idea of importance sampling~\cite{kirillov2020pointrend} by selecting stem-aware samples from the unlabeled data. In particular, considering that the unlabeled data are organized into $9$ domains, we select the top $500$ samples with the most proportion of stem pixels in each domain conditioned on the pseudo-labeled masks, resulting in $4500$ images in total. 
The stem-aware sample selection can maximize the presence of the stem class to the model, which somehow mitigates the problem of class imbalance.

The loss functions used follow the Mask2Former~\cite{he2022masked}. Given a wheat image $x_i$ and its ground-truth segmentation mask $y_i=\{(y_i^k,c_i^k)\}_{k\in[1,n]}$ encoded by binary masks $y_i^k$'s and class indices $c_i^k$'s. For each sample, the model predicts $K$ candidate masks $\{(\hat{y}_i^k, \hat{c}_i^k)\}_{1\leq k \leq K }$. After bipartite matching, a matching $(k, m_k)$ is obtained between the predictions and ground truths, where $m_k$ is the index of the predicted mask matched to the $k$-th ground truth mask. The single-sample supervised loss $\mathcal{L}_s^i$ thus can be defined by
\begin{equation}
\begin{aligned}
    \mathcal{L}_s^i=&\frac{1}{n}\sum_{k=1}^{n} \ell_{ce}(\hat{y}_i^{m_k}, y_i^k) +\lambda_{dice} \ell_{dice}(\hat{y}_i^{m_k}, y_i^k) \\
    &+\lambda_{c}\ell_{c}(\hat{y}_i^{m_k}, y_i^k)
\end{aligned}\,,
\end{equation}
where $\ell_{ce}$ is the cross-entropy loss, $\ell_{dice}$ is the dice loss, $\ell_{c}$ is the binary cross-entropy, and $\lambda_{dice}$ and $\lambda_{c}$ are hyperparameters used to control the contribution of different losses. 
The unsupervised loss $\mathcal{L}_u$ w.r.t. the unlabeled data is identical to the supervised one $\mathcal{L}_s$, except that the ground truth labels are replaced by the pseudo masks generated from the teacher model. By removing the superscript $i$ for $\mathcal{L}_s^i$, the final guided distillation takes the form
\begin{equation}
    \mathcal{L} = \mathcal{L}_s+\lambda_u\mathcal{L}_u\,,
\end{equation}
where 
$\lambda_u$ weights the importance of the 
unsupervised loss. Note that, guided distillation only uses $99$ labeled images and the selected $4500$ unlabeled images.


\subsection{Zoom In and Segment Twice}
Small- and fine-object visual perception has 
been 
long-standing hard cases for 
semantic segmentation. 
In our case, the \texttt{stem} class is identified 
to be the bottleneck due to its not only small-proportion but also fine-structure characteristics. After the stage-two training, we have achieved fairly good performance on the leaderboard, but not competitive enough to secure the first place. Hence, in the rest time of the competition, we start to seek other training-free solutions to improve the performance of the \texttt{stem} class further. 
Inspired by slicing-aided hyper inference~\cite{akyon2022slicing} which has shown strong performance in detecting small objects, we 
adapt this idea to wheat segmentation and develop a test-time scaling strategy called \textit{zoom in and segment twice}. This strategy is initially inspired by an surprising observation that a significant performance improvement can be achieved by simply testing on a zoomed-in version of image. By scaling the image resolution from $512\times512$ to $768\times768$, the performance improves from $0.74$ to $0.76$, which suggests the \texttt{stem} class can benefit from high-resolution inference. 

Formally, given a scaling ratio $\sigma \geq 1$, we define a $\tt{ZoomIn}$ operator such that $I^{\sigma}={\tt{ZoomIn}}(I, \sigma)$, which
zooms in the input image $I$ from its original size $(H,W)$ 
to $I^\sigma$ a larger size $(\sigma H, \sigma W)$. During inference, given a sliding window size $k=(k_h \times k_w)$ with a stride $t=(t_h,t_w)$, where the subscript $h$ and $w$ indicate the height and width components, we further define a $\tt{Slide}$ operator such that 
$
\{I_{i,j}^{\sigma}\}={\tt{Slide}}(I^{\sigma}, k, t) \,,
$
which slides the image $I^{\sigma}$, crop it into sub-window $\{I_{i,j}^{\sigma}\}$, where $\{I_{i,j}^{\sigma}\}=\{I^{\sigma}[i:i+k_h,j:j+k_w]~|~i=0,t_h,...,\sigma H-k_h; j=0,t_w,...,\sigma W-k_w \}$.

For each sliding window, we have the prediction logit for each window $L_{i,j}^s={\tt{Infer}}(I_{i,j}^{\sigma})$, where ${\tt{Infer}}$ means the model inference process. We then apply zero padding to match each logit to the original input resolution and average all the logits. During logit averaging, a count map $C\in \mathbb{R}^{\sigma H\times \sigma W}$ is recorded to 
indicate the number of times each pixel covered by the sliding windows. The final aggregated logit $L^{s}$ is obtained by dividing the summed logits by the count map as
\begin{equation}
     L^{\sigma} 
     =\frac{1}{C}\cdot \sum {\tt Pad}({\tt Infer}({\tt Slide}({\tt ZoomIn}(I,\sigma)), k, t)\,,
\end{equation}
where ${\tt Pad}(\cdot)$ is the zero padding operator. 

The `zoom in' strategy can be easily extended to multi-scale scenarios, which implements the `segment twice' idea.
Given a set of multi-scale ratio $S=\{\sigma_0,\sigma_1,...,\sigma_n\}$. For each ratio, we apply the same single-scale inference process above to obtain $L^{\sigma_i}$.
Then, we can simply average the predicted logits across different scales to obtain the final logits
\begin{equation}
    L=\frac{\sum_{\sigma_i \in S}{\tt{DownSample}}(L^{\sigma_i}, \sigma_i)}{|S|}\,,
\end{equation}
where ${\tt{DownSample}(\cdot, \sigma)}$ means downsampling the input with the ratio $\sigma$.

\begin{table}[!t]
    \centering
    \renewcommand{\arraystretch}{1.5}
    \addtolength{\tabcolsep}{-1pt}
    \scalebox{0.86}{
    \begin{tabular}{@{}lcccc@{}}
    \toprule
        \textbf{Method} & \textbf{SAPA} & \textbf{guided distillation} & \textbf{test time scaling} & \textbf{mIoU} \\
    \hline
        baseline & & & & $0.7284$  \\
    \hline
        & \ding{51} & & & 0.7291 \\
        & \ding{51} & \ding{51} &  & 0.7432 \\
        & \ding{51} & \ding{51} & \ding{51} & 0.7704 \\
    \bottomrule
    \end{tabular}}
    \caption{\textbf{Ablation of different stages during the development phase}.}
    \label{tab:ablation}
\end{table}
\section{Results and Discussions}
\begin{figure*}[!t]
    \centering
    \includegraphics[width=\linewidth]{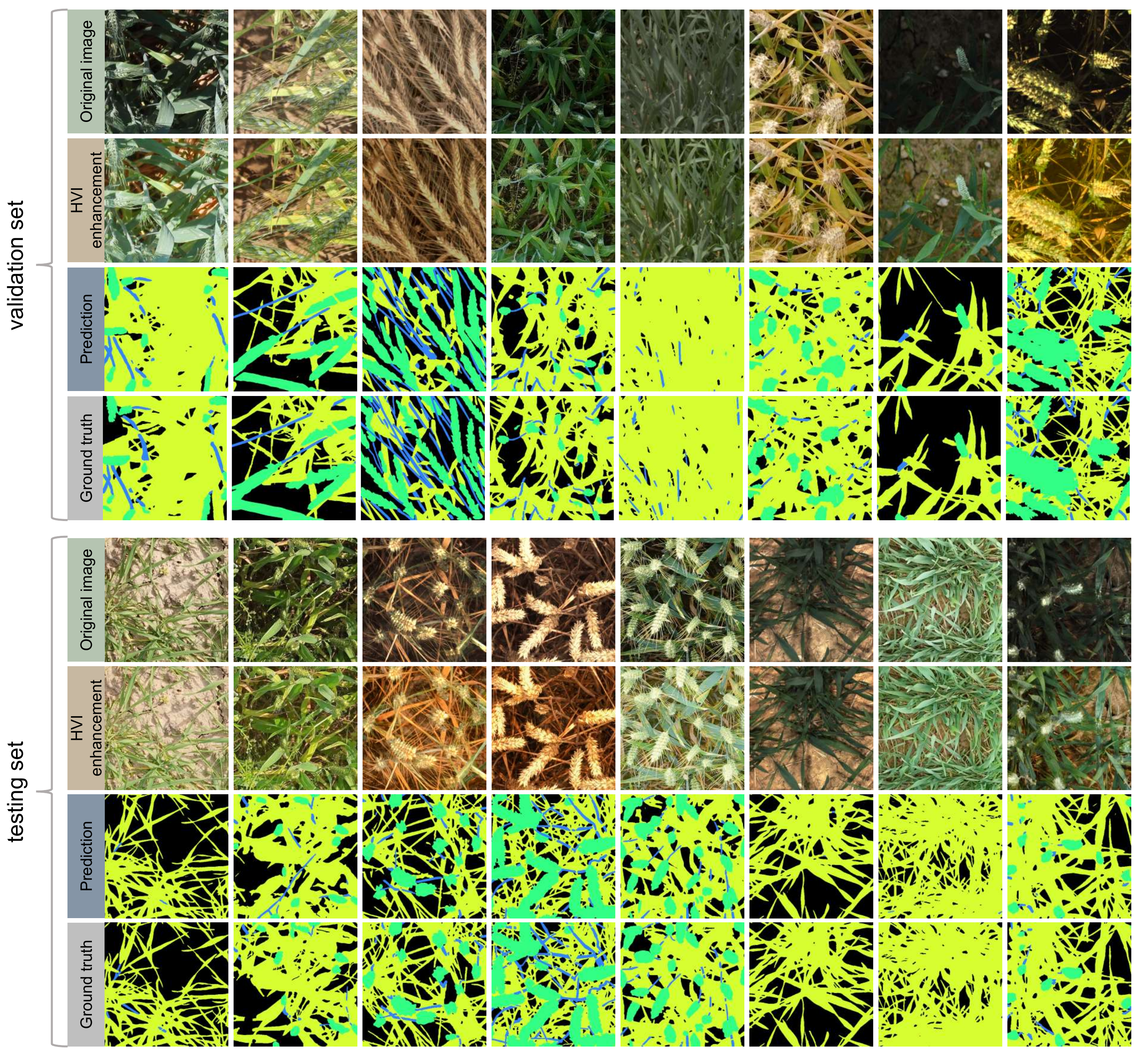}
    \caption{\textbf{Visualization of our model on validation and testing set}.}
    \label{fig:val_test_vis}
\end{figure*}
Here we present our results and discussions. Before that, we first introduce the evaluation metric used and our implementation details.

\subsection{Evaluation Metric}
The evaluation metric 
follows the standard semantic segmentation metric, that is, the mean Intersection over Union (mIoU), which is defined by
\begin{equation}
    \text{mIoU}=\frac{1}{N}\sum_{i=0}^N\frac{\text{TP}_i}{\text{TP}_i+\text{FP}_i+\text{FN}_i}\,,
\end{equation}
where $N$ is the number of classes. TP, FP, and FN denote the number of true positives, false positives, and false negatives for each class, respectively.

\subsection{Implementation Details}
We use the ImageNet-1K pretrained weight of BEiTv2-Large from the official repository, which has first trained for $1600$ epochs in a self-supervised manner and then finetuned for another 50 epochs following supervised learning. During our supervised learning stage for training a segmentation model, the model is trained for $20$K iterations, with a batch size of $16$. The base learning rate is set to $0.0001$, and the backbone learning rate is $0.01$. We use the ploy LRScheduler and set the warm up iterations to $0$. 
The first stage 
takes approximately 5 hours to complete. In the guided distillation stage, we set the total iterations to $90$K, with $20$K iterations for guided burn-in stage and apply early stop strategy. This stage requires about three days of training time. For the loss weights, we set $\lambda_{dice}=5$, $\lambda_{c}=5$, and $\lambda_u=2$. The EMA decay rate $\alpha$ is set to $0.9996$. In the testing stage, we use scaling ratios $\S=\{1.0, 1.5, 2.0, 2.5, 3.0, 3.5\}$ to scale the input image and apply sliding inference for each image. All experiments are conducted on the Ubuntu 20.04 system and on a workstation with four $48$ GB RTX A6000 GPUs, two $10$-core Intel Xeon Silver 4210R CPUs, and $256$ GB RAM.

\begin{table}[!t]
    \centering
    \begin{tabular}{@{}lc@{}}
    \toprule
       \textbf{Testing Scale}  &  \textbf{Leaderboard Score}\\
    \hline
        $512$ & $0.74$ \\
        $768$ & $0.76$ \\
        $512,768,1024,1280,1536,1792$ & $0.77$ \\
    \bottomrule
    \end{tabular}
    \caption{\textbf{Scores with different test-time scaling configurations}. Through the strategy of \textit{zoom in and segment twice}, we achieved a significant improvement. While extend our startegy to multi-scales, we can improve the performance further.}
    \label{tab:test_time_scale}
\end{table}

\begin{table}[!t]
    \centering
    \scalebox{0.87}{
    \begin{tabular}{@{}lcc@{}}
    \toprule
        \textbf{Team Name} & \textbf{Development Phase} & \textbf{Testing Phase} \\
    \hline
        HUST\_TinySmart (Ours) & $\textbf{0.77}$ & $\textbf{0.75}$  \\
        ZR Yang\&JH Jiang & $0.75$ & $0.71$  \\
        good & $0.75$ & - \\
        tapu1996 & $0.74$ & n/a \\
        pcccl & $0.74$ & $0.7$ \\
        willer & $0.74$ & - \\
        asuka & $0.74$ & $0.69$ \\
        nncckkuu & - & $0.68$ \\
        enchanter & $0.71$ & $0.66$ \\
        zaorui\_njfu & $0.71$ & $0.65$ \\
        Dafang Zou & $0.71$ & - \\
        tiezhuma & $0.7$ & $0.64$ \\
        phasheen & $0.69$ & - \\
        jaykk & $0.69$ & - \\
        xmba15 & $0.69$ & $0.63$ \\
        dsvolkov & $0.66$ & $0.62$ \\
        zengyangche & $0.67$ & $0.61$ \\
        perrychen & $0.64$ & - \\
        HZAU AISLE Group & $0.63$ & $0.57$ \\
        loannis Droutsas & $0.57$ & $0.54$ \\ 
        
    \bottomrule
    \end{tabular}}
    \caption{\textbf{Official leaderboard scores}. Top 20 teams from the development phase only (as of June 21, 2025). The scores are rounded mIoU.}
    \label{tab:leaderboard_score}
\end{table}

\subsection{Main Results}
The effect of our different stages in the development phase can
be found in Table~\ref{tab:ablation}. 
Results justify that each our design choice makes a difference. Surprisingly, the test time scaling stage yields the most significant improvement, improving the stage-two model by 0.0272 scores. A plausible explanation is that the model can better sense the detail of the stem class when zooming in the image. 
We also test the effect of different scaling ratios. The scores can be found in Table~\ref{tab:test_time_scale}. It can be observed that the single-scale zoom-in operation brings the most significant improvement. While our final solution uses the multi-scale inference strategy (segment twice), it introduces significant computational cost. 

The final competition results of the development and testing phases are shown in Table~\ref{tab:leaderboard_score}. Through our three improvements, we finally achieve a score of $0.7704$ in the development phase and $0.7468$ in the testing phase. Our solution ranks the first place on both the development and testing leaderboard. Remarkably, it exhibits the least performance drop in the testing phase, indicating strong generalization of our solution. 
In our view, this is mainly attributed to guided distillation, which enhances the training robustness of the model. Although our three improvements seem to be simple, we believe the key to our success is to follow the philosophy of the devil is in the detail and minority, that is, the \texttt{stem} class. The qualitative results of our final model are shown in Fig.~\ref{fig:val_test_vis}. 

\section{Conclusion}
This report presents our first place solution to the MLCAS 2025 GWFSS Challenge. Due to the significant illumination variations, we resort to the HVI color space for ease of analysis of segmentation results. 
Our approach is based on three insights observed from our baseline. By integrating the SAPA dynamic upsampling module into the ViT-Adapter, the ability of the model to capture fine details and tackle object boundaries is enhanced. Through guided distillation, we effectively leverage the unlabeled data, leading to improved model performance and enhanced robustness. Finally, through the strategy of \textit{zoom in and segment twice}, we further improve the performance of the model during the test time. 
We hope our solution can provide insights to future studies, especially on how to identify the bottleneck of the problem.
\section*{Acknowledgement}

This work is jointly supported by the Hubei Provincial Natural Science Foundation of China under Grant No. 2024AFB566 and by the HUST Undergraduate Natural Science Foundation under Grant No. 62500034.

{
    \small
    \bibliographystyle{ieeenat_fullname}
    \bibliography{main}
}

\end{document}